\documentclass{article}

\usepackage{arxiv}

\usepackage[utf8]{inputenc} % allow utf-8 input
\usepackage[T1]{fontenc}    % use 8-bit T1 fonts
\usepackage{hyperref}       % hyperlinks
\usepackage{url}            % simple URL typesetting
\usepackage{booktabs}       % professional-quality tables
\usepackage{amsfonts}       % blackboard math symbols
\usepackage{nicefrac}       % compact symbols for 1/2, etc.
\usepackage{microtype}      % microtypography
\usepackage{lipsum}		% Can be removed after putting your text content
\usepackage{graphicx}
\usepackage{natbib}
\usepackage{doi}
\usepackage{subcaption}
\usepackage{amsmath}

\title{$L^4$-Norm Weight Adjustments for Converted Spiking Neural Networks}

\author{{Jason M.~Allred\thanks{JMA performed this work while at Purdue University.}} \\
	Department of Computer Science and Electrical Engineering\\
	Brigham Young University--Idaho\\
	Rexburg, ID \\
	\texttt{allredjas@byui.edu} \\
	\And
	{Kaushik Roy} \\
	School of Electrical and Computer Engineering\\
	Purdue University\\
	West Lafayette, IN \\
	\texttt{kaushik@purdue.edu} \\	
}

\date{}

\renewcommand{\shorttitle}{$L^4$-Norm Weight Adjustments for Converted SNNs}

\hypersetup{
pdftitle={L4-Norm Weight Adjustments for Converted Spiking Neural Networks},
}

\begin{document}
\maketitle

\begin{abstract}
Spiking Neural Networks (SNNs) are being explored for their potential energy efficiency benefits due to sparse, event-driven computation.
Non-spiking artificial neural networks are typically trained with stochastic gradient descent using backpropagation.
The calculation of true gradients for backpropagation in spiking neural networks is impeded by the non-differentiable firing events of spiking neurons.
On the other hand, using approximate gradients is effective, but computationally expensive over many time steps. 
One common technique, then, for training a spiking neural network is to train a topologically-equivalent non-spiking network, and then convert it to an spiking network, replacing real-valued inputs with proportionally rate-encoded Poisson spike trains.
Converted SNNs function sufficiently well because the mean pre-firing membrane potential of a spiking neuron is proportional to the dot product of the input rate vector and the neuron weight vector, similar to the functionality of a non-spiking network.
However, this conversion only considers the mean and not the temporal variance of the membrane potential.
As the standard deviation of the pre-firing membrane potential is proportional to the $L^4$-norm of the neuron weight vector, we propose a weight adjustment based on the $L^4$-norm during the conversion process in order to improve classification accuracy of the converted network.
\end{abstract}

% keywords can be removed

\section{Introduction}

Spiking Neural Networks (SNNs) are a category of artificial neural networks used in machine learning that compute with event-driven internal signals triggered by the temporal `spikes' or firing events of spiking neurons.
SNNs differ from non-spiking feed-forward networks in several ways: (1) they have a built-in temporal component internal to each neuron creating a recurrent dependency over time; (2) there is some form of temporal encoding for the input, output and internal data signals; (3) they can operate with event-driven computation; (4) their non-linear activation functions are generally non-differentiable; and (5) the outputs of individual neurons are non-continuous, often being represented as binary--active (spike) or inactive (no spike).
Yet, much of the initial exploration of SNNs has followed the same paths as non-spiking networks, often ignoring some or all of these differences.
The neuron and synapse models and input temporal encoding must also be considered to more effectively and efficiently transfer the success of non-spiking networks to SNNs.

An efficient way of training spiking neural networks has been to train a topologically homogeneous non-spiking network and then converting the weights over to a spiking network, replacing ReLUs with integrate-and-fire neurons or with leaky-integrate-and-fire (LIF) neurons and then balancing the weights with the thresholds, as in  \cite{diehl2015thresholdbalancing, sengupta2019deepspikingconversion}.
One issue with this method is that SNNs operate in the temporal domain, and this training process completely ignores any temporal variance.
By accounting for the temporal variance, the conversion process can be improved.

We have identified a connection between the standard deviation of the pre-firing membrane of a spiking neuron and the $L^4$-norm of its weight vector.
Using this relationship, we are able to make adjustments to the weight vectors such that the internal spike rates more closely align with the activation values of the non-spiking neurons from which they were trained, improving classification accuracy of the converted SNN.

In this work, we first examine the statistical membrane potential distribution of spiking neurons (Section \ref{sec_distribution}) before using that information to  perform the $L^4$-norm weight adjustments during conversion (Section \ref{sec_l4}). We then detail our experimental methodology (Section \ref{sec_method}) and present our simulation results (Section \ref{sec_results}) which quantify the $L^4$-norm adjustment benefit.

\section{Mean and Variance of Spiking Neuron Membrane Potentials}
\label{sec_distribution}

As the membrane potential is a function over a stochastic process, it is a time-dependent random variable $V(t)$. In \cite{allred2020dopaminergic}, specifically Section 2.3.4 and Supplemental Section 1 therein, we included a derivation of the mean and variance of the pre-firing membrane potential of an LIF neuron receiving a Poisson rate-encoded spike train input, which may be viewed as a shot-noise process (\cite{PhysRevE.63.031902}). The distribution itself may be approximated as lognormal, being the summation of the exponential decay of values drawn from a uniform distribution. 

For a given neuron $j$, the first moment or mean of its pre-firing membrane potential $V_j(t)$ may be written as:

\begin{equation}
    E[V_j(t)] = \tau_{mem}(\vec{w_j}\cdot\vec{\lambda})(1-e^{-t/\tau_{mem}})
    \label{EQ_combined_shot_noise_mean}
\end{equation}

where $t$ is the time since the start of the spike train or the time since the last membrane potential reset, $\tau_{mem}$ is the decay time constant of the membrane potential, $\vec{w_j}$ is the weight vector of neuron $j$, and $\vec{\lambda}$ is the input rate vector.
In steady-state this converges to: $\tau_{mem}(\vec{w_j}\cdot\vec{\lambda})$, which is the basic input-weight vector dot product essential to the non-spiking to spiking conversion process.

Continuing on to the second moment, the variance of $V_j(t)$ is given by:

\begin{equation}
    Var(V_j(t)) = \frac{1}{2}\tau_{mem}(\vec{\lambda}\cdot\vec{w_j}^{\circ 2})(1-e^{-2t/\tau_{mem}})
    \label{eq_final_variance}
\end{equation}
where $\vec{w_j}^{\circ 2}$ represents the \textit{Hadamard square}, or element-wise square, of the weight vector.
This equation is important for discussions next in Section \ref{sec_l4}.

\section{$L^4$-norm Weight Adjustments for SNNs Conversions}
\label{sec_l4}

In this section, we demonstrate why the variance of the membrane potential should be considered, then connect the standard deviation of the membrane potential to the $L^4$-norm of its weight vector before proposing the conversion weight adjustments.

\subsection{Why account for membrane potential variance?}
\label{sec_why_variance}

\begin{figure}[p]
\begin{center}
\includegraphics[width=4.5in]{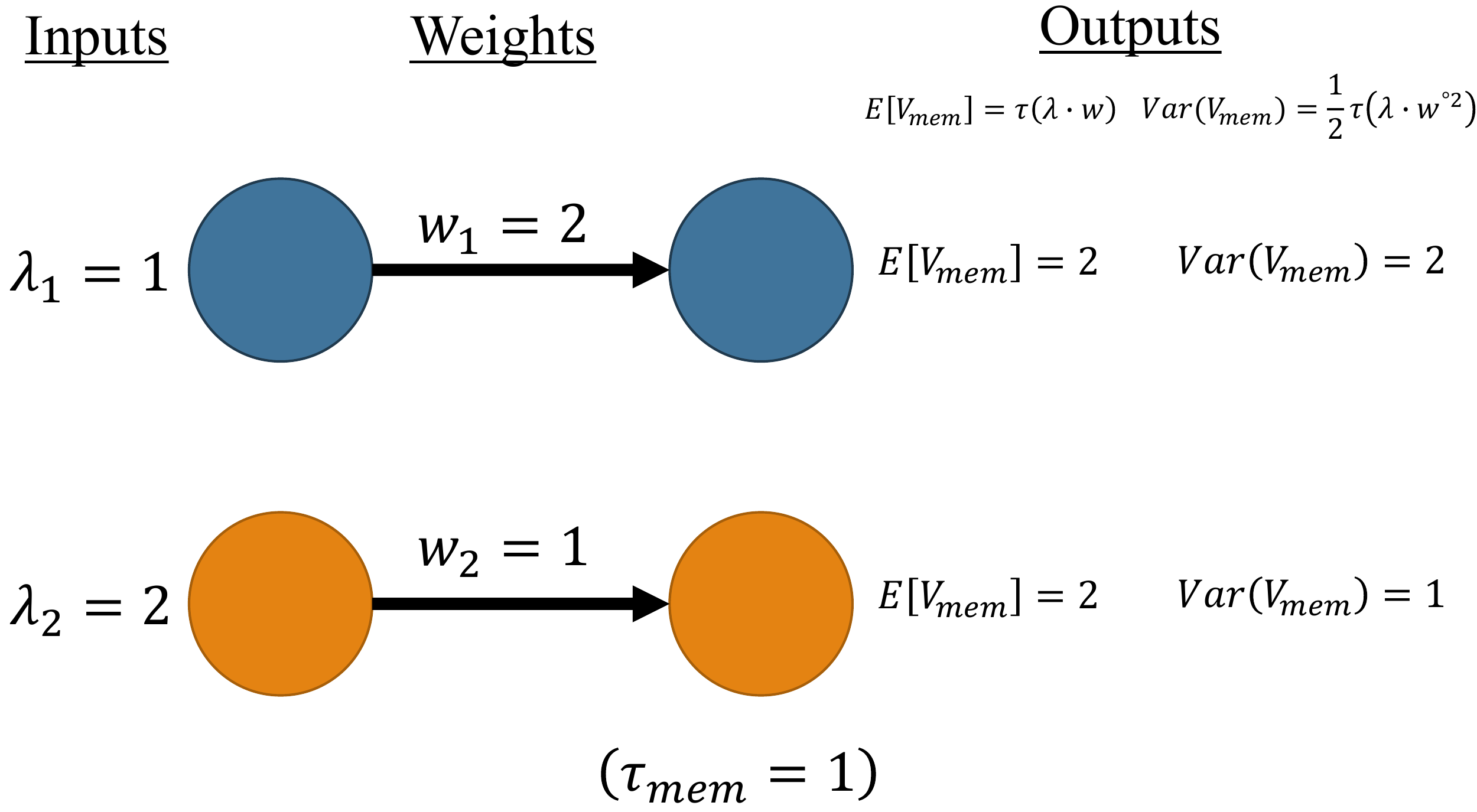}
\end{center}
\caption{A simple example of the same mean membrane potential, but different variances.}
\label{fig_l4_1D_example}
\end{figure}

\begin{figure}[p]
\begin{center}
\includegraphics[width=6.5in]{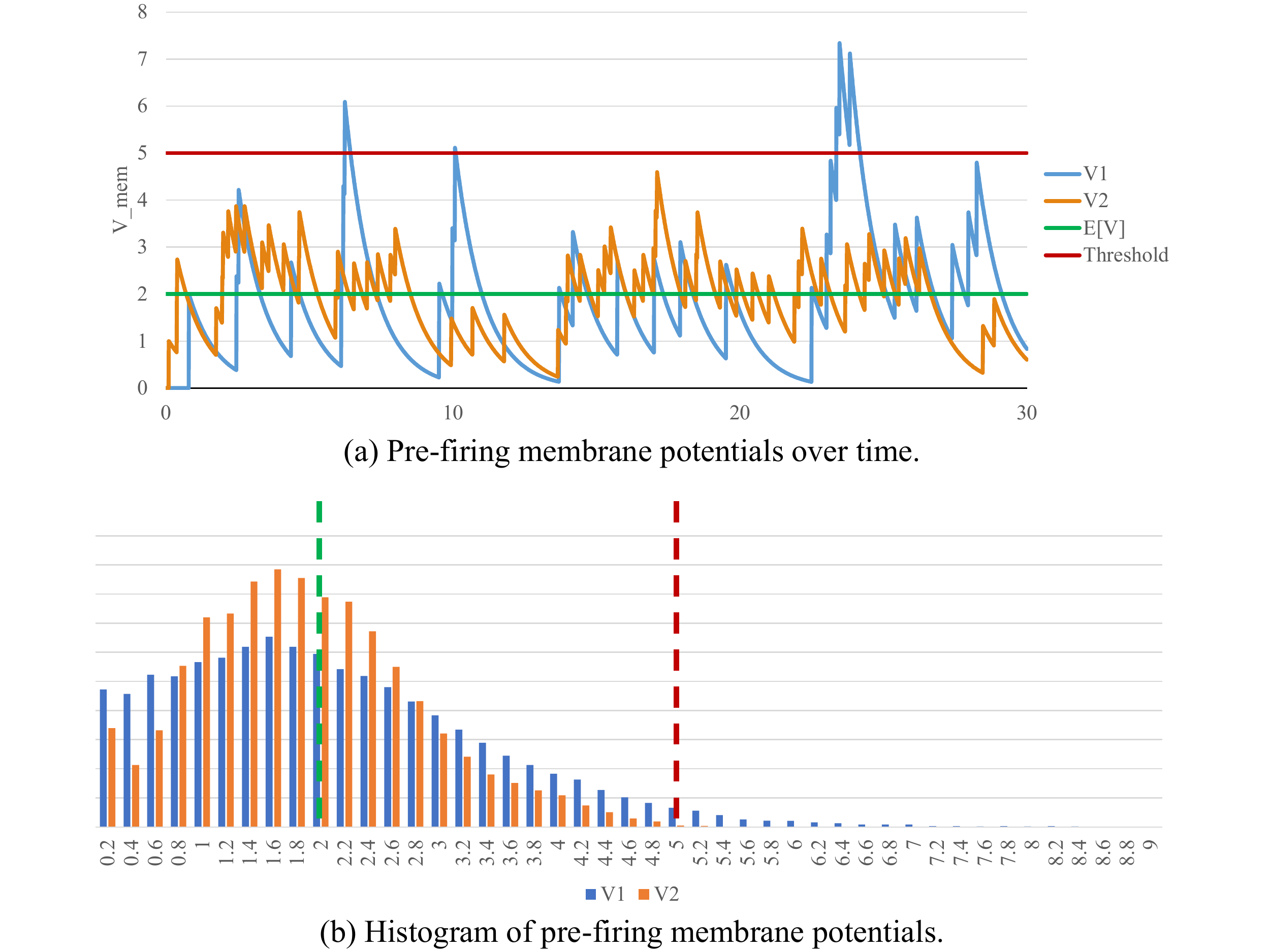}
\end{center}
\caption{The temporal responses and distributions of the examples in Figure \ref{fig_l4_1D_example}.}
\label{fig_l4_1D_distribution}
\end{figure}

Although the mean pre-firing membrane potential of an LIF neuron is proportional to the dot product of the input vector and the weight vector, we cannot assume that a neuron that has a larger mean pre-firing membrane potential will have a higher spiking probability.
To illustrate this claim, consider two membrane potential distributions which have equivalent means but different variances, for example two neurons each with a single input and $\tau_{mem} = 1$: the first with input rate $\lambda_1 = 1$ and synaptic weight $w_1 = 2$, and the second with input rate $\lambda_2 = 2$ and synaptic weight $w_2 = 1$. Both have the same mean membrane potential, $E[v_{mem1}] = E[v_{mem2}] = 2$; however the first neuron has a larger variance, $Var(v_{mem1}) = \frac{1}{2}(1)(1)(2)^2 = 2$; and $Var(v_{mem2}) = \frac{1}{2}(1)(2)(1)^2 = 1$. (See Figure \ref{fig_l4_1D_example}.) Given a certain threshold higher than the mean, the distribution with the larger variance will have a higher probability of surpassing the threshold than the distribution with a smaller variance (see Figure \ref{fig_l4_1D_distribution}).

\begin{figure}
\begin{center}
\includegraphics[width=5.5in]{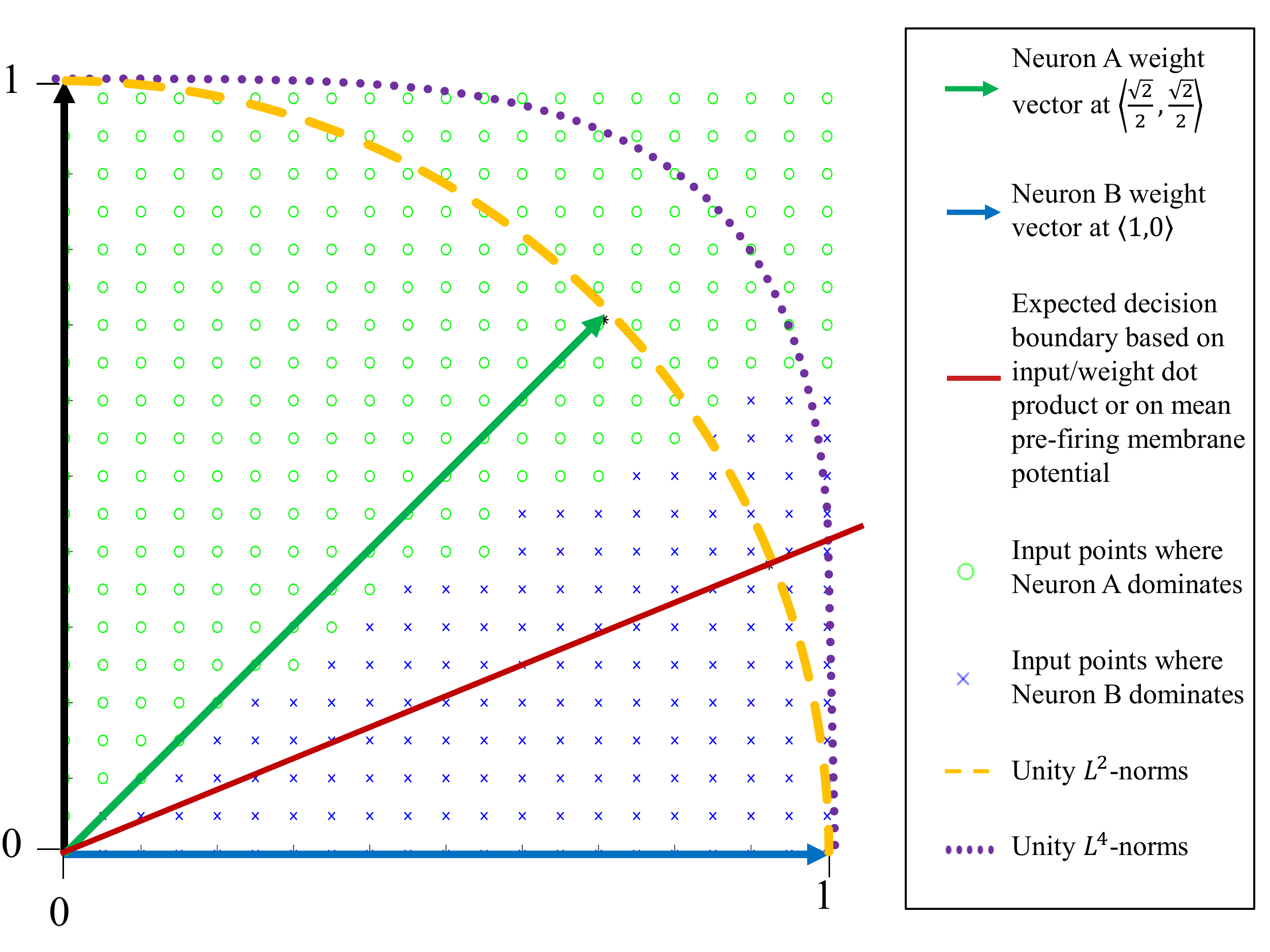}
\end{center}
\caption{A two-dimensional example of neurons with the same weight magnitude (same $L^2$-norm), showing the input points at which each dominates in a spiking domain.}
\label{fig_l4_2D_example}
\end{figure}

The mismatch between means and variances isn't solely based on weight vector magnitudes, either. Consider another simple example expanded into another dimension, with two inputs and weights per neuron (see Figure \ref{fig_l4_2D_example}). Based on the input/weight vector dot products, we would expect the decision boundary between these two neurons to be at the angular midpoint. However, in the spiking domain Neuron B dominates in much of the input space that is angularly closer to Neuron A. Neuron B was able to dominate more of the input space than expected because its membrane potential at those points in the input space had a larger variance than did the membrane potential of Neuron A. The difference in variance at those points is because while both neurons have weight vectors with the same magnitude, using a euclidean or $L^2$-norm, the Hadamard or element-wise squares of their respective weight vectors do not have the same magnitude.

\subsection{Relation of $v_{mem}$ standard deviation to the weight $L^4$-norm}

We know from Equation \ref{eq_final_variance} that the variance of the pre-firing membrane potential is proportional to the dot product of the input and the Hadamard (or element-wise) square of the weight vector, written as $w^{\circ2}$.
Therefore, the variance is proportional to the $L^2$-norm, or euclidean magnitude of $w^{\circ2}$, which can be rewritten as:

\begin{align}
    \nonumber
    |w^{\circ 2}|_{L^2} &= \sqrt{(w_1^2)^2 + (w_2^2)^2 + (w_3^2)^2 + ... + (w_n^2)^2} \\
    \nonumber
    &= \sqrt{w_1^4 + w_2^4 + w_3^4 + ... + w_n^4} \\
    \nonumber
    &= \left(\sqrt[4]{w_1^4 + w_2^4 + w_3^4 + ... + w_n^4}\right)^2 \\
    &= (|w|_{L^4})^2 \\
    \nonumber
    \label{eq_final_L4}
\end{align}

And since the standard deviation is the square root of the variance, the standard deviation of the membrane potential becomes directly proportional to the $L^4$-norm of the weight vector. Going back to the example in Figure \ref{fig_l4_2D_example}, we now see that the standard deviation of the membrane potential of Neuron B was larger than the membrane potential of Neuron A at the points in question because Neuron B's weight vector had a larger $L^4$-norm than did Neuron A's, despite them having equivalent $L^2$-norms.

\subsection{Performing the $L^4$-norm adjustments}

For a given $L^2$-norm, a weight vector that is closer to an axis will have a larger $L^4$ norm. These are weight vectors that are more sparse or have a few larger individual components rather than many fairly equivalent individual components. For example, the vectors $<~\frac{\sqrt{2}}{2}$,~$\frac{\sqrt{2}}{2}~>$ and $<~1,~0~>$ have the same $L^2$-norm, but their respective $L^4$ norms are $\approx 0.84$ and $1$.

Current conversion methods from trained non-spiking ANNs to SNNs assume that a neuron's output spike rate is proportional to the dot product of the input rate vector and the neuron's weight vector.
Failing to account for the wider reach of neurons with a larger $L^4$-norm can cause non-ideal conversions.
We proposed scaling a neuron's threshold, or equivalently its weight magnitude, by some function of the $L^4$-norm of the pre-conversion weight vector so that systems that operate on a dot-product assumption incur less of a ``conversion penalty.''

As a reminder, the mathematical modeling of the statistical distribution of the neuron pre-firing membrane potential that we have been using was built on the \textit{Poisson} rate-encoding of the input.
We note here that after the input layer, the spike trains may no longer be considered Poisson point processes, as the potential has a transient state, meaning that the probability of a neuron that is within the network firing is dependent on how recently it has previously fired and reset its potential.
This observation would imply that the following adjustments are more effective at the first layer of the network. Our simulations verify that, and so the following adjustments are applied to the first layer only.

\section{Adjustment Methodology}
\label{sec_method}

For clarity, we will begin by discussing how an individual neuron's firing threshold $v_{th}$ should be scaled. In practice, we instead scale the weights by the inverse amount, which has the exact same functionality but still allows us to maintain a single threshold voltage for all neurons in the layer. The reason for calculating the threshold change instead of the weight magnitude change is so that the neuron's pre-firing membrane potential distribution remains consistent throughout the calculation process.

\subsection{Scaling by the weight vector's $L^4$-norm}

As discussed above, because training in a non-spiking domain is based on the dot product of the input and weight vectors, the mean pre-firing membrane potential of neurons in the converted network will be proportional to that dot product.
For more accurate use of those parameters in the converted SNN, we instead want those computed values to be reflected in each neuron's respective probability of having its potential reach its firing threshold.
We propose scaling each neuron's threshold to a new value $\widehat{v}_{th_j}$ so that the number of \textit{standard deviations} it is over the neuron's mean membrane potential is proportional to the difference between the original threshold $v_{th_0}$ and the mean.

\begin{equation}
    \frac{\widehat{v}_{th_j} - \mu_{V_j(t)}}{\sigma_{V_j(t)}} \propto v_{th_0} - \mu_{V_j(t)}
    \label{eq_L4_reach}
\end{equation}

In other words, we want the reach of the threshold over the mean to be scaled by the standard deviation, which is proportional to the $L^4$-norm:

\begin{equation}
    \frac{\widehat{v}_{th_j} - \mu_{V_j(t)}}{v_{th_0} - \mu_{V_j(t)}} \propto  \sigma_{V_j(t)} \propto |\vec{w_j}|_{L^4}
    \label{eq_L4_reach_2}
\end{equation}

To keep the parameters in the same range as training, we scale it with the average $L^4$-norm of the neurons in that layer, $\mu_{L^4}$:

\begin{equation}
    \nonumber
    \frac{\widehat{v}_{th_j} - \mu_{V_j(t)}}{v_{th_0} - \mu_{V_j(t)}} = \frac{|\vec{w_j}|_{L^4}}{\mu_{L^4}} 
\end{equation}   

which allows us to calculate the new threshold for each neuron:
    
\begin{equation} 
    \widehat{v}_{th_j} = \mu_{V_j(t)} + \frac{|\vec{w_j}|_{L^4}}{\mu_{L^4}}(v_{th_0} - \mu_{V_j(t)})
    \label{eq_L4_reach_3}
\end{equation}

The mean membrane potential is input-dependent, and cannot be calculated only with the weight values. We sample it by running a forward pass on a single batch of the training set through only the first layer to measure the mean membrane potential of each neuron. This extra single batch of inference need not add much computation time to the conversion because the single-batch inference is already performed during the threshold balancing step of the conversion, and we simply record the mean potential within the first layer during that inference.

Then, as Equation \ref{eq_L4_reach_3} gives us the new threshold, we instead use its proportion to the original threshold inversely as the $L^4$-norm weight adjustment scalar, $s_{L^4}$, as mentioned at the start of this subsection:

\begin{equation} 
    s_{L^4} = 1/(\frac{\widehat{v}_{th}}{v_{th}})
    \label{eq_L4_reach_4}
\end{equation}

\subsection{Smoothing ratio, $r_s$}

\begin{figure}
\begin{center}
\includegraphics[width=5.5in]{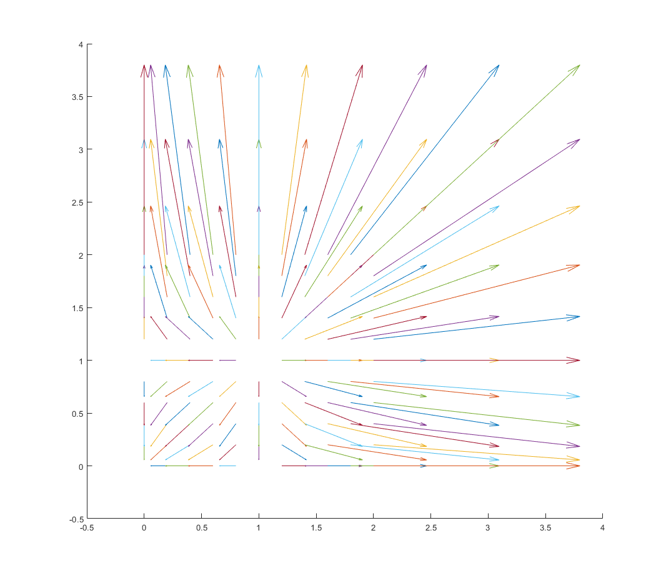}
\end{center}
\caption{The Hadamard angle drift. In this figure, each vector represents the transformation of performing an element-wise square--the shown start point being an original vector endpoint and the shown endpoint being the endpoint of the element-wise square of the original vector.}
\label{fig_hadamard}
\end{figure}

The variance is also input dependent. Although the magnitude of a given input doesn't have an affect of the conversion penalty--since it scales all the potentials of all competing neurons equivalently--the angle of the input plays a role.
The variance is proportional to the cosine of the angle between the input vector $\vec{\lambda}$ and the Hadamard square of the weight vector $\vec{w_j}^{\circ 2}$.
Figure \ref{fig_hadamard} shows a 2-d plot of how performing the element-wise square on a vector changes not only its magnitude, but also its direction.

Vectors that are along an axis or are along the ones vector do not change their angle from the origin, but every other vector drifts away from the ones vector toward the axes.
This means that for a given angle between $\vec{\lambda}$ and $\vec{w_j}$, the corresponding angle between $\vec{\lambda}$ and $\vec{w_j}^{\circ 2}$ will either be smaller or larger, depending on if the input vector is located in the direction of the drift or away from it, respectively.

\begin{figure}
\begin{center}
\includegraphics[width=6.5in]{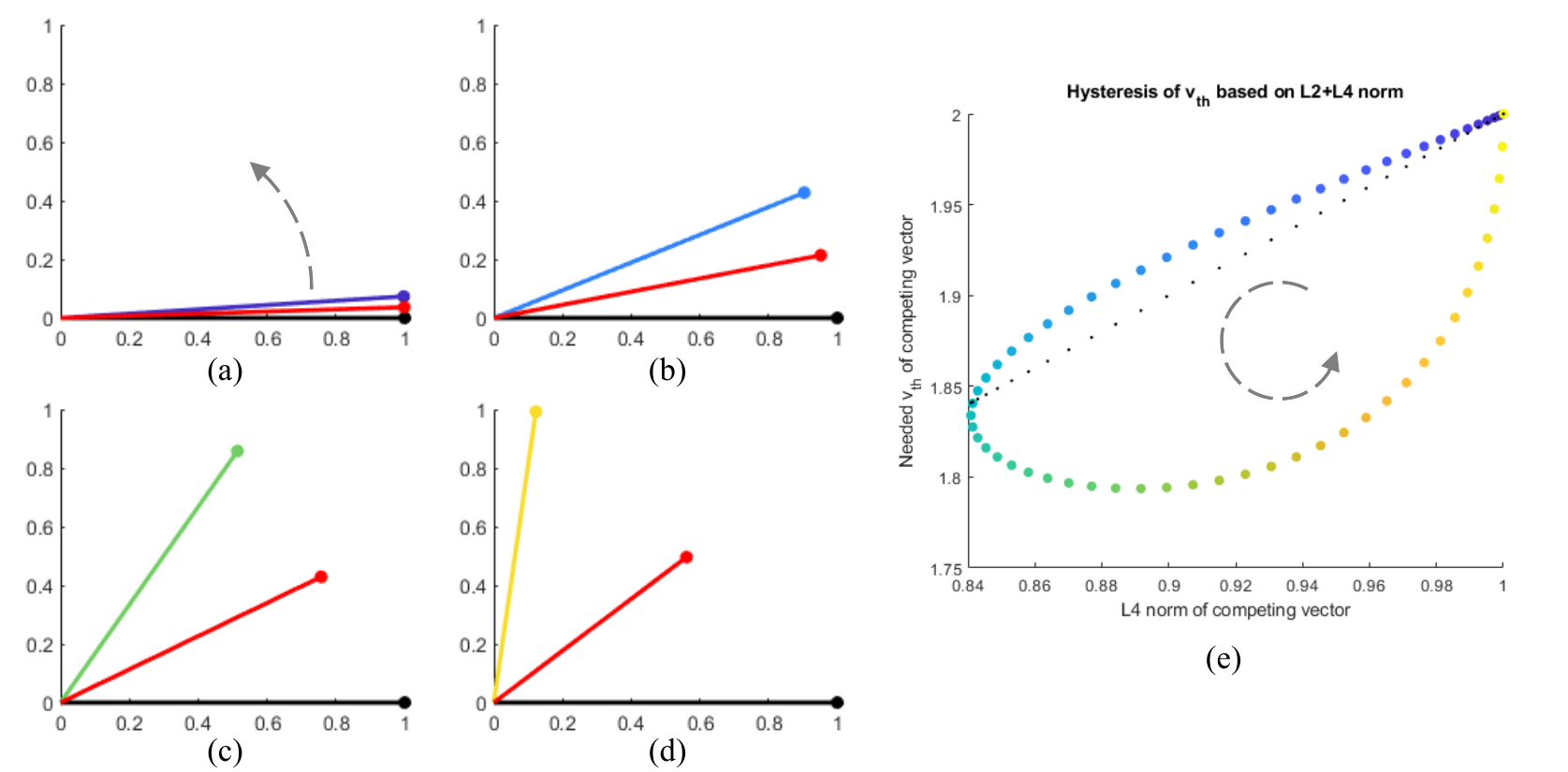}
\end{center}
\caption{A 2-d simulation of the needed $v_{th}$ of a sweeping weight vector (purple to blue to green to yellow) that is competing against a static weight vector (black) at <1,0> when presented with the midpoint input vector (red). The dotted gray arrow indicates the direction of the sweep. Subfigures (a), (b), (c), and (d) are snapshots of the sweep from an angle of zero to $\pi$/2. Subfigure (e) shows the corresponding color coded $v_{th}$ of the sweeping neuron in order for it to achieve the same spiking activity as the constant neuron, whose $v_{th}$ remains static at 2. The smaller black dots represent the $v_{th}$ that would correspond to a mean + variance of the $L^2$ + $L^4$ norms. ($\tau=1$)}
\label{fig_hysteresis}
\end{figure}

This creates a hysteresis (see Figure \ref{fig_hysteresis}), meaning that there cannot be a single $v_{th}$ for a neuron that is input-independent and completely correct for the temporal variance.
What this means is that in some cases, the $L^4$-norm correction will overshoot, and potentially move to a portion of the loss space that increases loss rather than decreases loss.

We minimize this effect by smoothing the $L^4$-norm variance estimations with a weighted average of all the $L^4$-norms in the layer, where the layer average $\mu_{L^4}$ is weighted by $r_s$, a hyperparameter, as follows:

\begin{equation}
    |w_i|_{\widehat{L^4}} = \frac{(|w_i|_{L^4} + r_s \mu_{L^4})}{1+r_s}
\end{equation}

Then $|w_i|_{\widehat{L^4}}$ is used in place of $|w_i|_{L^4}$ in the weight adjustment calculation of \ref{eq_L4_reach_3}.

Further, there is the additional approximation of the mean potentials that are profiled from a single batch and averaged over those inputs and across windows of convolution rather than input-dependent for each sample. We additionally apply the smoothing ratio, then, to determine how much of the correction to apply. The larger the smoothing ratio $r_s$, the smaller the weight adjustment. 

\begin{equation}
    \hat{s}_{L^4} = \frac{s_{L^4} + r_s}{1 + r_s} = \frac{s_{L^4} - 1}{1 + r_s} + 1
\end{equation}

Finally, after scaling the weights in the first layer by $\hat{s}_{L^4}$, we calculate the overall change in the average weight magnitude of the layer and re-scale the whole layer to match the original average weight magnitude so that the spike rate remains on-par with the baseline.

\subsection{Threshold balancing}

In traditional non-spiking to spiking conversion, a threshold must be assigned to each layer. Because we are changing the distribution of the membrane potentials and squashing the outliers, the same threshold that was optimal for the baseline may not be the same threshold that is optimal for the $L^4$-norm adjusted networks. (Recall that the adjustment didn't actually change the threshold, but instead the weights inversely.) Therefore, for a more fair comparison, we perform the same threshold voltage sweep for both the baseline and the adjusted networks and choose the best for each.

\section{Adjustment Results}
\label{sec_results}

We tested Cifar-10 on both VGG-5 and RESNET-20 as well as Cifar-100 on VGG-11 using both soft-resent and hard-reset neurons\footnote{These pre-print results represent preliminary simulations. Additional runs are underway to increase statistical confidence in the measured improvements. As such, the data in this section may be hereafter updated.}. We also tested five different leak scalars for each network, performing the threshold and smoothing ratio hyperparameter sweep on each using the training data. (Note that these values show the leak scalar at each time step rather than the corresponding leak time constant $\tau$. For these figures, 1 means no leak, and lower values mean faster leak.) The ANN to SNN conversion followed the methodology, code, and some of the same pre-trained networks as in \cite{Rathi2020Enabling}.

\subsection{Cifar-10 on VGG-5}

\begin{figure}
\begin{center}
\includegraphics[width=6.5in]{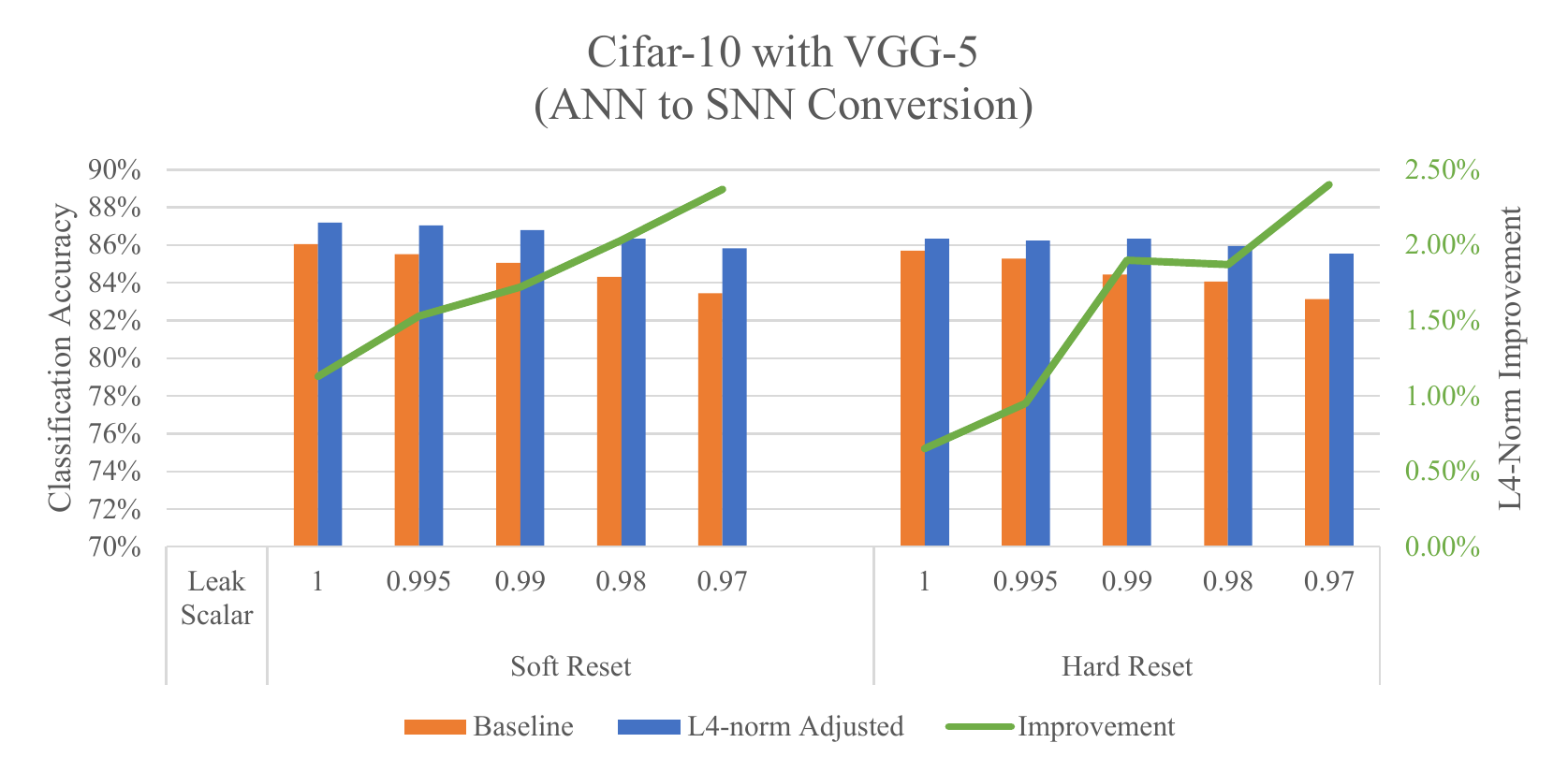}
\end{center}
\caption{Classification accuracy of an SNN converted from a pre-trained non-spiking ANN with the VGG-5 network model on the Cifar-10 dataset, showing the improvement by adjusting weights according to the $L^4$-norm.}
\label{fig:l4_results_vgg5}
\end{figure}

When VGG-5 is trained on Cifar-10 in a non-spiking ANN, the ANN accuracy is 89.32\%. We convert that pre-trained network to a spiking network operating with 75 time steps per inference. Fig. \ref{fig:l4_results_vgg5} shows the accuracy when converting to a spiking network for both the baseline conversion and the $L^4$-norm adjusted conversion.  The $L^4$-norm adjusted networks outperform the baseline in all scenarios and regains some of the accuracy lost during conversion without needing any additional training in the spiking domain (which is costly).

As expected, for all networks the accuracy drops when the leak is faster and also drops with the hard reset, in both cases because of information loss.
However, it is still beneficial to explore these scenarios. Leaky neurons allow for identifying temporal patterns that can be erased by non-leaky neurons, and there are also emerging analog devices that have intrinsic parasitic sub-threshold leak. Additionally, hard resets can sometimes be cheaper to implement in hardware, as a value simply needs to be discharged to ground rather than performing a subtraction operation.
Note that the $L^4$-norm improvement increases as the leak gets faster. This is because the temporal stochasticity and variance plays a larger roll in the spiking activity when there is more leak.

However, we still get an improvement even in the no-leak scenario, which we had originally not expected. In the no-leak scenario, the traditional conversion dropped 3.27 percentage points to 86.05\% while the $L^4$-norm adjusted network was able to regain 1.13 of those lost percentage points. The reason this was unexpected is that because without leak, there is no steady-sate; every positively-bound membrane potential will continue to increase at an expected rate proportional to the input-weight dot product and eventually fire. In fact, mathematically, the expected time to fire--and thus the spiking activity--of a no-leak neuron will always be proportional to the desired input-weight dot product, no matter the variance in that rate. The benefit here, therefore, must be that that by stabilizing the variance of that firing rate, it reduces the temporal noise in the spike rates for the subsequent layers.

\subsection{Cifar-10 on RESNET-20}

\begin{figure}
\begin{center}
\includegraphics[width=6.5in]{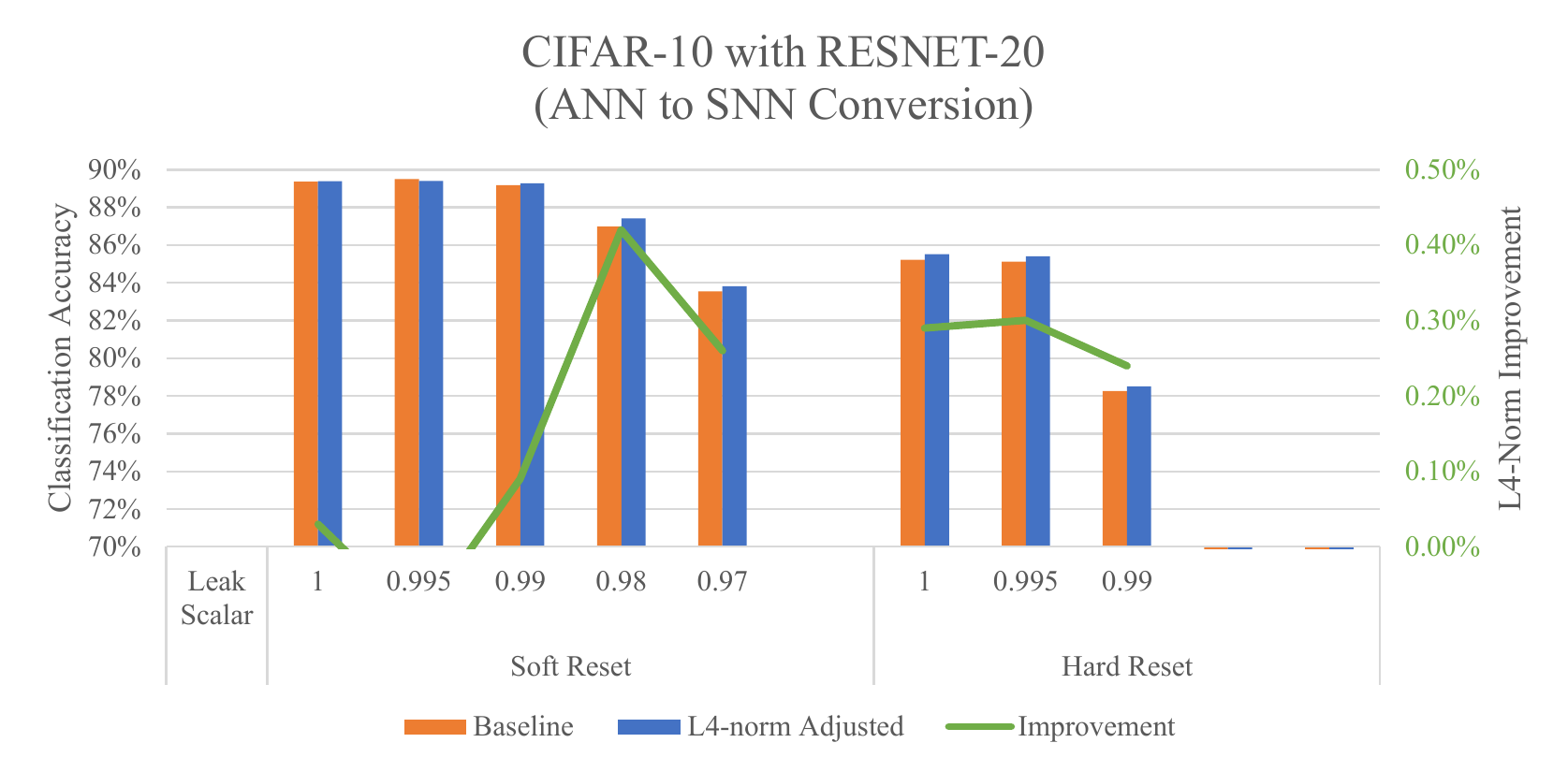}
\end{center}
\caption{Classification accuracy of an SNN converted from a pre-trained non-spiking ANN with the ResNet-20 network model on the Cifar-10 dataset, showing the improvement by adjusting weights according to the $L^4$-norm.}
\label{fig:l4_results_resnet20}
\end{figure}

When RESNET-20 is trained on Cifar-10 in a non-spiking ANN, the ANN accuracy is 92.79\%. We convert that pre-trained network to a spiking network operating with 250 time steps per inference. Fig. \ref{fig:l4_results_resnet20} shows the accuracy when converting to a spiking network for both the baseline conversion and the $L^4$-norm adjusted conversion. (Note that because of information loss, RESNET-20 failed to generalize for the faster two leaks in the hard reset scenario for both the baseline and the $L^4$-norm adjusted networks, and are not shown.)

In contrast to the smaller network VGG-5, RESNET-20 did not see a benefit in the no- and low-leak, soft-reset scenarios by performing the $L^4$-norm adjustment. We suspect that because RESNET-20 is a larger network, the de-noising of the internal spike rates was less effective because the redundancy was able to handle the noise. Yet, the benefit still exists in the faster leak scenarios, verifying that even with a larger network we need to correct for the temporal variance affecting the ability to reach the firing threshold.

We further note that in the hard reset scenarios, the $L^4$-norm adjustment \textit{did} provide a benefit even in the no- and slow-leak scenarios. This benefit may be because the $L^4$-norm adjustment reduces the overshoot of high-varying potentials over the threshold, reducing the associated information loss. Thus there are three areas in which adjusting for the $L^4$-norm provides a benefit: (1) in all networks with sufficient leak, by accounting for the gap between the mean and threshold in proportion to the standard deviation; (2) in no-leak networks that are smaller, or less redundant by reducing the spike rate noise past the first layer; and (3) in networks with hard-reset neurons, even without leak and with redundancy, by accounting for the statistical overshoot past the threshold, reducing information loss.

\subsection{Cifar-100 on VGG-11}

\begin{figure}
\begin{center}
\includegraphics[width=6.5in]{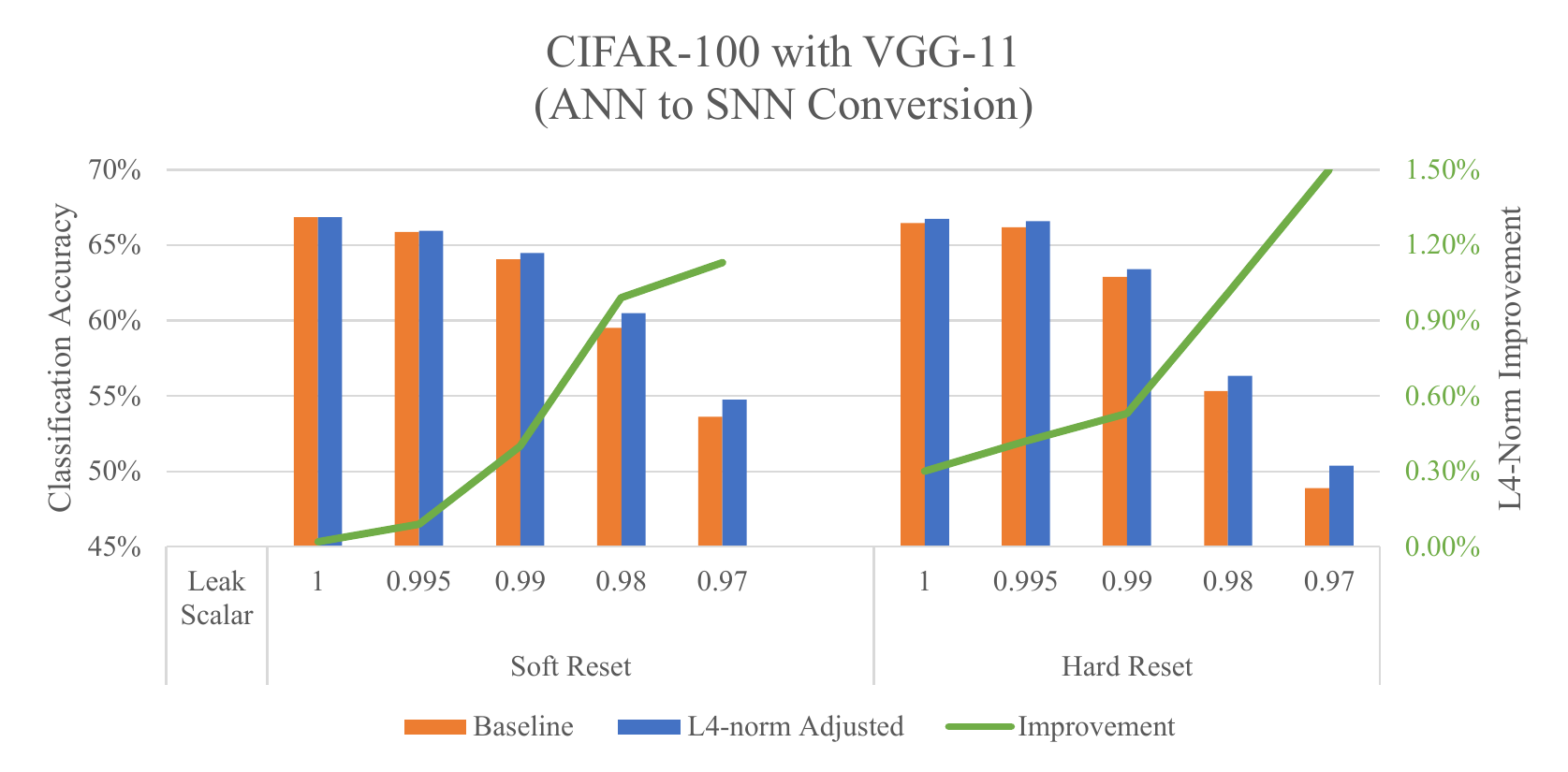}
\end{center}
\caption{Classification accuracy of an SNN converted from a pre-trained non-spiking ANN with the VGG-11 network model on the Cifar-100 dataset, showing the improvement by adjusting weights according to the $L^4$-norm.}
\label{fig:l4_results_vgg11}
\end{figure}

When VGG-11 is trained on Cifar-100 in a non-spiking ANN, the ANN accuracy is 71.21\%. We convert that pre-trained network to a spiking network operating with 125 time steps per inference. Fig. \ref{fig:l4_results_vgg11} shows the accuracy when converting to a spiking network for both the baseline conversion and the $L^4$-norm adjusted conversion.

With this network and dataset, we see similar areas of improvement as with RESNET-20. While there is little to no benefit with soft reset and no- or slow- leak, there is still the expected benefit at faster leaks and for hard reset. The overall improvement for VGG-11 is higher than that of RESNET-20, likely because CIFAR-100 is a more complicated dataset and so the corrections play a bigger role.

\section{Conclusion}

The temporal variance of a spiking neuron's pre-firing membrane potential should be considered when transferring learned weights from a non-spiking neural network to a spiking neural network. We have shown that the standard deviation of the pre-firing membrane potential of a spiking neuron is proportional to the $L^4$-norm of its weight vector. That relationship allows us to perform an adjustment to the converted SNN thresholds, or equivalently to the converted SNN weight magnitudes. This adjustment, when smoothed to account for input dependence, can improve the accuracy of a converted SNN without any expensive training in the spiking domain.

\section*{Conflict of Interest Statement}

The authors declare that the research was conducted in the absence of any commercial or financial relationships that could be construed as a potential conflict of interest.

\section*{Acknowledgements}
We gratefully acknowledge useful discussions with Professor Jonathon Peterson of Purdue University regarding the derivations used in Section \ref{sec_distribution}.

\section*{Funding}
This work was supported in part by C-BRIC, a JUMP center sponsored by the Semiconductor Research Corporation and DARPA, and by the National Science Foundation, Intel Corporation, and the Vannevar Bush Fellowship.

\section*{Data Availability Statement}
The CIFAR dataset used in this study can be found at \href{https://www.cs.toronto.edu/~kriz/cifar.html}{[https://www.cs.toronto.edu/~kriz/cifar.html]}.

\bibliographystyle{unsrtnat}
\bibliography{references}

\end{document}